\pdfoutput=1

\documentclass[11pt]{article}

\usepackage[]{acl}

\usepackage{times}
\usepackage{latexsym}
\usepackage{hyperref}       
\usepackage{url}            
\usepackage{booktabs}       
\usepackage{amsfonts}       
\usepackage{nicefrac}       
\usepackage{xcolor}         
\usepackage{enumitem}

\usepackage{array}
\usepackage{multirow}
\usepackage{graphicx}
\usepackage{soul}
\usepackage[ruled]{algorithm2e}
\usepackage{color}
\usepackage{wrapfig}
\usepackage{amsmath}
\usepackage{cleveref}
\usepackage{ulem}
\usepackage[T1]{fontenc}
\usepackage[utf8]{inputenc}
\usepackage{microtype}
\usepackage{inconsolata}
\usepackage{subcaption}
\usepackage{pifont}
\usepackage{ulem}
\usepackage{hyperref}

\crefformat{section}{\S#2#1#3} 
\crefformat{subsection}{\S#2#1#3}
\crefformat{subsubsection}{\S#2#1#3}

\definecolor{myorange}{RGB}{255, 190, 122}
\definecolor{myblue}{RGB}{166, 191, 231}
\definecolor{myblue2}{RGB}{6, 70, 173}

\newcommand{\kilm}{{KILM}}
\newcommand{\continuedtraining}{{continued pre-training}}

\title{\kilm: Knowledge Injection into Encoder-Decoder Language Models}

 \author{Yan Xu\textsuperscript{{\rm 1,2}}\thanks{\  \ Work done in part while Yan was an intern at Amazon Alexa AI.}\ , Mahdi Namazifar\textsuperscript{{\rm 1}},  Devamanyu Hazarika\textsuperscript{{\rm 1}}, 
\textbf{Aishwarya Padmakumar\textsuperscript{{\rm 1}}}, 
\\ \textbf{Yang Liu\textsuperscript{{\rm 1}}, Dilek Hakkani-T\"ur\textsuperscript{{\rm 1}}} \\
\textsuperscript{\rm 1}Amazon Alexa AI \\
\textsuperscript{\rm 2}Hong Kong University of Science and Technology\\
\texttt{yxucb@connect.ust.hk}, \texttt{mahdinam@amazon.com}, \texttt{dvhaz@amazon.com}\\
\texttt{padmakua@amazon.com}, \texttt{yangliud@amazon.com}, \texttt{hakkanit@amazon.com}
}

\begin{document}
\maketitle

\begin{abstract}
Large pre-trained language models (PLMs) have been shown to retain implicit knowledge within their parameters. To enhance this implicit knowledge, we propose Knowledge Injection into Language Models (KILM), a novel approach that injects entity-related knowledge into encoder-decoder PLMs, via a generative knowledge infilling objective through continued pre-training. This is done without architectural modifications to the PLMs or adding additional parameters. 
Experimental results over a suite of knowledge-intensive tasks spanning numerous datasets show that KILM enables models to retain more knowledge and hallucinate less, while preserving their original performance on general NLU and NLG tasks. KILM also demonstrates improved zero-shot performances on tasks such as entity disambiguation, outperforming state-of-the-art models having 30x more parameters. \footnote{The code is available at \href{https://github.com/alexa/kilm}{https://github.com/alexa/kilm}.}

\end{abstract}

\section{Introduction}

Large pre-trained language models (PLMs)~\citep{radford2019language,lewis2020bart,raffel2020exploring} have achieved great success across all NLP tasks. However, recent studies also reveal that PLMs are susceptible to memorizing the pre-training corpora rather than capturing the knowledge within them~\citep{niven2019probing,talmor2020olmpics,yasunaga2022linkbert,li2022instilling}. Particularly for generation tasks, PLMs are notorious for hallucinating text that is factually incorrect or hard to verify~\citep{logan2019barack,sun2020ernie,lin2020commongen,longpre2021entity}.
To address these issues, one approach is to retrieve relevant knowledge and integrate it explicitly with PLMs~\citep{he2020bert,liu2021kg}. 
Another direction is incorporating the additional knowledge sources into the pre-training step~\citep{zhang2019ernie,xiong2019pretrained,liu2021knowledge,wang2021kepler}.
While the former suffers from the issue of falling back on the models themselves without retrieved information~\citep{krishna2021hurdles},
knowledge-focused pre-training can be complementary to those methods~\citep{longpre2021entity} and shows its advantage on generalization.

\begin{figure*}[t]
  \centering
  \includegraphics[width=.95\linewidth]{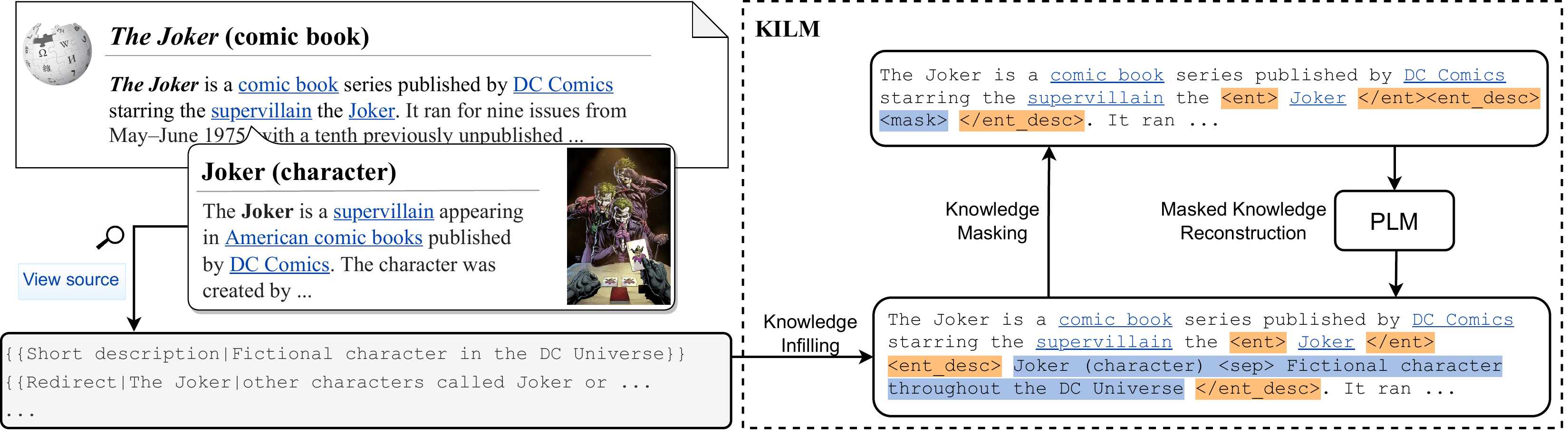} 
  \caption{The illustration of the proposed KILM technique for injecting knowledge into PLMs. In the given example, the mention, \textit{Joker}, is linked to the page of Wikipedia entity \textit{Joker (character)}. While the figure only shows knowledge infilling, knowledge masking, and masked knowledge reconstruction steps, the proposed method is combined with the original pre-training objectives of PLMs for continued pre-training.}
  \label{fig:object}
\end{figure*}

In this paper, we propose an approach for injecting knowledge into encoder-decoder PLMs, such as BART, as a \continuedtraining~process. We refer to it as \textit{Knowledge Injection into Language Models} (\kilm). Instead of introducing additional parameters to PLMs or modifying the model architectures to incorporate additional knowledge, \kilm~infills knowledge sentences by adopting a novel knowledge infilling objective that includes a knowledge reconstruction step in addition to the original pre-training objectives of BART. 

The aim of KILM is to teach PLMs additional content about concepts and entities that they encounter in a given context, so that the models are able to ground an entity mention with additional information and ``describe'' what that entity is (see \Cref{fig:object}). It should be emphasized that in this process, the context is especially important for cases when an entity mention can refer to multiple entities, e.g., \textit{Titanic} which can refer to the \textit{British ship} or to the \textit{1997 movie}. We utilize the \textit{short descriptions} of entities in Wikipedia which comprise of entity definitions as the knowledge source (\Cref{sec:preliminary}).
Although there are existing works leveraging similar knowledge for PLM enhancement, they ignore the relationship among entities, contexts, and entity-centric knowledge, and restrict their applications to NLU tasks. In contrast, we propose a distinct structure (\Cref{sec:kilm}) to augment Wikipedia articles with short descriptions of the entity mentions in the context, thus model this essential relationship, so as to force PLMs to learn the correlation among entities and contexts, and differentiate between the entities with similar surface forms during continued pre-training. 
With recent work that highlights the need for explicit grounding for PLMs to truly understand text~\cite{merrill2021provable}, we posit that KILM takes a step in that direction.

The proposed structure for knowledge infilling in KILM is further leveraged as a structured prompt in downstream tasks (see \Cref{sec:kn_tasks}). 
We demonstrate better knowledge retention with KILM in 
zero-shot for entity disambiguation and appositive generation tasks, showing the effectiveness of the proposed method. Even without the distinct structure, we also find that BART with \kilm~outperforms BART on QA tasks and is less prone to hallucination on tasks such as knowledge-grounded response generation. 
As mentioned earlier, KILM relies on \continuedtraining~of PLMs, which presents the possibility of catastrophic forgetting of original skills of the PLM. We mitigate this by retaining the original training objectives of BART during the \continuedtraining~stage. We empirically verify that our proposed objective does not degrade the general language modeling ability of the PLM, nor affect the fluency of these models for natural language generation (NLG) tasks. 
Although we focus on short descriptions of entities as the knowledge source for \kilm, other forms of knowledge can also be used, which we leave for future exploration.

We summarize our contributions as follows:
    
    \noindent(1) We propose a novel approach, KILM, to leverage Wikipedia annotations in pre-training of PLMs. We inject knowledge into BART, solely through \continuedtraining, with no change in the architecture of the PLMs. KILM enables entity-based knowledge injection with knowledge in natural-language form. KILM's distinct structure also offers a direct way to probe the entity knowledge retained in pre-trained models.
    
    \noindent(2) We show that KILM enhances the performance of BART on knowledge-intensive tasks while maintaining its original performance on other downstream tasks. KILM demonstrates improved zero-shot performance on entity disambiguation task, outperforming state-of-the-art models having 30x more parameters.

\section{Related Work}

\paragraph{Knowledge-Enhanced LMs} To enhance PLMs' use of knowledge, a number of work has attempted to augment them with external knowledge sources, such as knowledge graphs (KGs)~\citep{yin2022survey}. 
Some recent work introduced additional non-parametric memories into the models~\citep{zhang2019ernie,rosset2020knowledge} to obtain entity embeddings and modified the model structures to accommodate extra information~\citep{yamada2020luke,wang2021k,wang2021kepler}, while others changed the masking schema with the additional information~\citep{sun2019ernie,wang2022effectively}, or converted the external KGs into natural language text as an additional pre-training corpus~\citep{xiong2019pretrained,zhou2020pre,liu2021knowledge,agarwal2021knowledge,li2022instilling}. 


\paragraph{Modeling with Text Linking and Enrichment} 

Our motivation bears similarity to \textit{text linking}~\citep{yasunaga2022linkbert,deng2021reasonbert,arora2022metadata} during pre-training and \textit{text enrichment}~\citep{elazar2021text}. Modeling the links between documents or metadata is motivated by the fact that PLMs, pre-trained on plain text, are not directly trained to capture inter-dependencies between documents.
The similarity between the above tasks and ours lies in the ways humans implicitly \textit{link} information when reading or generating language. However, the former tasks are restricted to relationships within the text, while our goal is to ground the concepts and entities to their related descriptions in encyclopedic resources.

\paragraph{Pre-training with Hypertext}
\label{sec:html_cm3}

Besides PLMs that are pre-trained with natural language corpora, HTLM~\citep{aghajanyan2021htlm} directly pre-trains simplified crawled HTML data based on BART models and CM3~\citep{aghajanyan2022cm3} extends HTLM into a multimodal setting with causal masked language modeling. 
The target of HTLM and CM3 is to better leverage the enormous web-scraped data source for pre-training. 
In contrast, our work aims to leverage hypertext to explore how to inject extra knowledge into PLMs with a custom-designed structure to furnish advantages to PLMs in performing knowledge-intensive tasks.

\section{Methodology}

Although \kilm~is model-agnostic and could be used for any PLM (more on this in \cref{sec:other-plm}), in this work, due to high computation costs, we focus on applying KILM to BART~\citep{lewis2020bart}.

\subsection{Preliminaries}
\label{sec:preliminary}
 
Wikipedia is a widely-used text corpus for LM pre-training. It is often processed as a collection of individual articles in the form of flat natural language text. However, due to the existence of hyperlinks in its text, Wikipedia is also a complex web of connected Wikipedia topics, also known as Wikipedia entities. These hyperlinks build connections between different Wikipedia entities and establish a rich source of information that is mostly ignored in current pre-training approaches. Moreover, most Wikipedia articles come with a \textit{short description} of the entity (topic) discussed in the article. These short descriptions provide definitions for Wikipedia entities. In this work, we take an initial step towards using these additional information within Wikipedia articles and utilizing ``short descriptions'' of entities for \continuedtraining~of PLMs. 
Note that 
the proposed approach could be expanded to \textbf{other annotated text corpora}.

\subsection{\kilm: Knowledge Injection into Language Models}
\label{sec:kilm}
We propose KILM, which extends the text-infilling objective to knowledge infilling objective through continued pre-training. 
KILM, as shown in \Cref{fig:object}, consists of three steps: (1) \textit{knowledge infilling}, (2) \textit{knowledge masking}, and (3) \textit{masked knowledge reconstruction}.

\paragraph{Knowledge Infilling}
As mentioned in \Cref{sec:preliminary}, in this work, we mainly focus on injecting PLMs with hyperlinks and entity descriptions as the entity-related knowledge into PLMs. 
Specifically, we process Wikipedia data such that entity mentions in Wikipedia articles (which are annotated by hyperlinks) are marked with a start-of-entity token \texttt{<ent>} and an end-of-entity token \texttt{</ent>}. Also, each entity mention is followed by an entity-related knowledge sentence marked with \texttt{<ent\_desc>} and \texttt{</ent\_desc>} as start- and end-of-description tokens. The inserted knowledge component (highlighted in blue in~\Cref{fig:object}) consists of the corresponding hyperlinked entity (which might be different from the entity's surface form in the text) and the entity's short description connected with the \texttt{<sep>} token, where the short description is obtained from a lookup table extracted from the Wikipedia dump.
We denote this knowledge infilling transformation as \textsc{KnInfill}.


\paragraph{Knowledge Masking}
The processed data is used for the \continuedtraining~of a PLM. During this step, we conduct knowledge masking transformation (denoted as \textsc{KnMask}) and the model is trained to reconstruct the whole inserted knowledge component from a single \texttt{<mask>} token with respect to the context. More specifically, assuming the $i$th token $t_i$ is a mention of an entity, the masked input sequence $\mathbf{X}$ and the output sequence $\mathbf{Y}$ can be denoted as:
\setlength{\abovedisplayskip}{2pt}
\setlength{\belowdisplayskip}{2pt}
\begin{align*}
    \begin{split}
        \mathbf{X} =& \{t_1, ..., t_{i-1}, \colorbox{myorange}{\texttt{\small<ent>}}, t_i, \colorbox{myorange}{\texttt{\small</ent>}, \texttt{\small<ent\_desc>}}, \\  
        & \colorbox{myblue}{\texttt{\small<mask>}}, \colorbox{myorange}{\texttt{\small</ent\_desc>}}, t_{i+1}\, ..., t_N\}, 
    \end{split}
    \\
    \begin{split}
        \mathbf{Y} =& \{t_1, ..., t_{i-1}, \colorbox{myorange}{\texttt{\small<ent>}}, t_i, \colorbox{myorange}{\texttt{\small</ent>}, \texttt{\small<ent\_desc>}}, \\ 
        & \colorbox{myblue}{$k_1$, ..., $k_L$}, \colorbox{myorange}{\texttt{\small</ent\_desc>}}, t_{i+1}\, ..., t_N\},
    \end{split}
\end{align*}
where $t_n$ represents the $n$th token of the original target sequence and $k_l$ represents the $l$th token in the knowledge sequence of length $L$.

\begin{table*}[t]
\centering
\resizebox{.95\linewidth}{!}{
\begin{tabular}{@{}lcccll@{}}
\toprule
\textbf{Task} & \textbf{\begin{tabular}[c]{@{}c@{}}Knowledge\\ type \end{tabular}} & \textbf{\begin{tabular}[c]{@{}c@{}}Task\\ adapation\end{tabular}} & \textbf{Input/Prompt} & \textbf{Target} \\ \midrule
\begin{tabular}[c]{@{}l@{}}Entity \\ Disambigua- \\tion\end{tabular} & entity & \ding{55} & \begin{tabular}[c]{@{}l@{}}\textbf{\textit{Context} $D$:} The Big Blue River is ... Driftwood White, \\\colorbox{myorange}{\texttt{<ent>}}Wabash\colorbox{myorange}{\texttt{</ent><ent\_desc>}}\colorbox{myblue}{\texttt{<mask>}} \\ \colorbox{myorange}{\texttt{</ent\_desc>}}, and ... \\ \textbf{\textit{Candidate} $S^1$:} Wabash River\texttt{<sep>}Tributary of the Ohio ...\\ \textbf{\textit{Candidate} $S^2$:} Wabash, Indiana\texttt{<sep>}Wabash is a city in ... \\ \end{tabular} & 
Wabash River \\ \midrule
\begin{tabular}[c]{@{}l@{}}Appositive \\Generation\end{tabular} & entity & \ding{55} & \begin{tabular}[c]{@{}l@{}}The game achieved the highest ... matchup between Larry \\ Bird and Spartans' point guard \colorbox{myorange}{\texttt{<ent>}} Magic Johnson \\ \colorbox{myorange}{\texttt{</ent><ent\_desc>}} \colorbox{myblue}{\texttt{<mask>}} \colorbox{myorange}{\texttt{</ent\_desc>}}.\end{tabular} & \begin{tabular}[c]{@{}l@{}}a rivalry that lasted \\ throughout their \\ professional careers\end{tabular} \\ \midrule
\begin{tabular}[c]{@{}l@{}}In-Context \\ Few-Shot QA\end{tabular} & factoid & \ding{55} & \begin{tabular}[c]{@{}l@{}}Question: What jobs did Ben Franklin do? Answer: Diplomat\\ Question: What did Ben Franklin invent? Answer: \colorbox{myblue}{\texttt{<mask>}}\end{tabular} & \begin{tabular}[c]{@{}l@{}}Lightning rod \end{tabular} \\ \midrule
KGRG & \begin{tabular}[c]{@{}c@{}}encyclopedia\end{tabular} & \ding{51} & \begin{tabular}[c]{@{}l@{}}
<speaker2>Ross was an American painter and television host. \\ <speaker1>That's cool. What else? \\ <speaker2>\end{tabular}& \begin{tabular}[c]{@{}l@{}}He created the show \\ "The Joy of Painting"\end{tabular} \\ \bottomrule
\end{tabular}
}
\caption{A summary of the knowledge-intensive tasks that are studied in this work. \textit{KGRG} is short for \textit{Knowledge Grounded Response Generation} task. Examples of input and target formats are provided above along with the task information. The definitions of the knowledge types are discussed in the corresponding sections in \Cref{sec:kn_tasks}.}
\label{tab:examples}
\end{table*}

\paragraph{Masked Knowledge Reconstruction}
The parameters $\theta$ of the PLM are optimized by a masked knowledge reconstruction loss:

\setlength{\abovedisplayskip}{-10pt}
\setlength{\belowdisplayskip}{2pt}
\begin{equation*}
    \mathcal{L}_{kn} = \mathbb{E}\left(\sum_{l=1}^{L}-\log\left( p\left(k_l|t_{1:(i+l+2)}, \mathbf{X}, \theta\right)\right)\right).
\end{equation*}

Since our goal is to inject entity-related knowledge without disrupting the function of the original BART as a general PLM, the masked knowledge reconstruction loss is combined with the original text infilling objective of BART during \continuedtraining.\footnote{The comparison between the text infilling and sentence permutation objectives shows the advantage of the former objective over the latter~\citep{lewis2020bart}, so we only preserve the text infilling objective for KILM to simplify the \continuedtraining~task.}
At training time, the model is optimized by minimizing the reconstruction loss over the whole target sequence instead of only the recovered masked spans. 
As a result, the training objectives force the model to learn to copy the tokens from the input sequences when the token is not a mask token during the pre-training process. This is to help the model recognize the inserted knowledge components in the training sequences and ensure the fluency of the PLM on NLG tasks. 
The weights of different objectives for loss are calculated based on the proportion of the corresponding spans across the entire sequence.
We summarize the proposed KILM algorithm in \Cref{sec:appx:alg}. 

The advantages of leveraging this structure for training are two-fold. First, this structure builds an alignment between the entity-related knowledge and the corresponding mention in the paragraphs. 
Second, the injected knowledge can be easily induced by probing the PLM with the structured prompts proposed for KILM (\Cref{sec:kn_tasks}).


\begin{table*}[t!]
\centering
\resizebox{.95\linewidth}{!}{%
\begin{tabular}{@{}lcccccccr@{}}
\toprule
\textbf{Models} & \textbf{AIDA} & \textbf{MSNBC} & \textbf{AQUAINT} & \textbf{ACE2004} & \textbf{CWEB} & \textbf{WIKI} & \textbf{Avg} & \textbf{Parameters} \\ \midrule
CM3-medium~\citep{aghajanyan2022cm3}$^\ddagger$ & 78.0 & 80.1 & 75.4 & 81.4 & 68.5 & 76.2 & 76.6 & 2,700M \\
CM3-large~\citep{aghajanyan2022cm3}$^\ddagger$ & 80.1 & 80.8 & 77.7 & 82.8 & \textbf{72.4} & 80.2 & 79.0 & 13,000M \\ \midrule
BART-base       & 33.8 & 57.6 & 44.6 & 37.8 & 36.4 & 46.1 & 42.7 & 139M \\
BART-base+Merge & 28.2 & 43.3 & 27.1 & 19.5 & 27.3 & 39.9 & 30.9 & 139M \\
BART-base+KILM (ours)  & 80.0 & 83.7 & 74.7 & 78.2 & 63.7 & 71.3 & 75.3 & 139M \\ \midrule
BART-large      & 34.4 & 58.8 & 42.3 & 38.9 & 36.9 & 46.5 & 43.0 & 406M \\
BART-large+KILM (ours) & 84.6 & 86.4 & 79.8 & 80.9 & 66.1 & 75.4 & 78.9 & 406M \\
BART-large+KILM$_{\text{DU}}$ (ours) & \textbf{86.2} & \textbf{87.8} & \textbf{84.3} & \textbf{83.7} & 68.4 & \textbf{79.9} & \textbf{81.7} & 406M \\ 
\bottomrule
\end{tabular}
}
\caption{InKB Micro F1 on zero-shot entity disambiguation tasks with candidates from \citet{le2018improving}. $^\ddagger$The results are from CM3 under the zero-shot setting.}
\label{tab:el}
\end{table*}

\section{Experiments}

We start by exploring the performance of BART+\kilm~on knowledge-intensive tasks (\Cref{sec:kn_tasks}). Later, we also demonstrate that \kilm~does not degrade the original language modeling skills of BART in both NLU and NLG benchmarks (\Cref{sec:gen_tasks}). 

\subsection{Pre-training Details}

\paragraph{Data}
\label{sec:data}


To extract the short descriptions and the hyperlinks from Wikipedia articles, we process a Wikipedia dump from scratch.\footnote{The Wikipedia dump is downloaded from \url{https://dumps.wikimedia.org/enwiki/}.} We assign the first sentence of the Wikipedia page as the short description if the ``short description'' attribute is missing in the raw data. 
We use the processed data by only leveraging the paragraphs from the summary sections of Wikipedia as our primary training corpus (denoted as \textit{primary setting}), while we also explore a \textit{data upscaling setting} where we use the entire Wikipedia articles.
We split the articles with document strides of 512 and consider one snippet as a data sample. We randomly select one entity from the paragraphs in each iteration for dynamic entity-centric knowledge injection.\footnote{We select different entities in each iteration.} After data pre-processing, we obtain a collection of 5.70 million data samples for the primary setting and 7.85 million data samples for the data upscaling setting from Wikipedia.
We split the corpus into a training set and a validation set with around 10k samples, for evaluation. In the following sections, KILM without a subscript indicates that it is conducted under the default primary setting, while KILM under data upscaling setting will be denoted as KILM$_{\text{DU}}$.
For pre-training in the primary setting, the model is continually trained for 7,000 steps, and for the data upscaling setting, the model is trained for 50,000 steps.\footnote{Most of our results are based on KILM in the  primary setting, and due to the computational resource cost, only for a subset of knowledge intensive tasks we also report the results for data upscaling setting too.} Refer to \Cref{sec:appx:pretrain} for details.


\paragraph{Baselines}

Besides the original BART, we also report on another BART-base baseline that is continue pre-trained on a merge of Wikipedia corpus and short descriptions for 7,000 steps (same number of steps as KILM) with only text infilling objective. The short descriptions are converted to general text based on the format: \textbf{\texttt{``<Entity> is <Short Desc>''}}. This model is denoted as BART-base+Merge. We demonstrate input and output formats of pre-training in \Cref{tab:io}. This baseline is introduced to separately evaluate the role of the distinct structure that is introduced in this work, as well as the additional training steps and data.

\subsection{Knowledge-Intensive Tasks}
\label{sec:kn_tasks}

First, we study the effectiveness of KILM on knowledge-intensive tasks~\citep{petroni2019language,roberts2020much,petroni2021kilt}. As shown in \Cref{tab:examples}, we evaluate BART+KILM on \textbf{\textit{entity disambiguation}} and \textbf{\textit{appositive generation}} tasks, which have similar objectives to the \continuedtraining~of KILM. 
We also evaluate if KILM can contribute to downstream tasks where the pre-training objective of KILM is not fully aligned with those of the downstream tasks. Specifically, We include 
\textbf{\textit{question answering (QA)}} and \textbf{\textit{knowledge grounded response generation (KGRG)}} tasks.

\paragraph{Zero-shot Entity Disambiguation}
\label{sec:ed}
The entity disambiguation task requires the model to link a mention $q$ to the correct entity, given a context $\mathbf{D}$ and several candidate entities. \textbf{Without fine-tuning}, we evaluate BART+KILM by picking the candidate with the lowest perplexity of generating short descriptions $\{\mathbf{S}^{i}\}_{i=1}^N$ using structured prompts among the candidate entities $\{\mathbf{E}^i\}_{i=1}^N$ in entity disambiguation datasets.\footnote{Note that the reference entities in this task come from Wikipedia, hence we can use the associated entity description for each reference entity.} It can be expressed as:

\setlength{\abovedisplayskip}{-13pt}
\setlength{\belowdisplayskip}{0pt}

\begin{align}
    \mathbf{X}_i &= \textsc{KnMask}(\textsc{KnInfill}(\mathbf{D}, q, \mathbf{S}^i)) \\
    E^{i^*} &= \mathop{\arg\max}\limits_{i} \sum_{t} \log p(s_t^i|\mathbf{X}_i, \theta).
\end{align}
We use the same datasets and candidate sets as those in \citet{le2018improving}.
InKB micro-F1 results are shown in \Cref{tab:el}, where CM3, a series of huge PLMs trained with multimodal hypertext (see \Cref{sec:html_cm3}), are tested in a zero-shot setting. We also included the performances of BART and BART-base+Merge for reference.\footnote{More details are included in \Cref{sec:ed2}.}
BART+KILM outperforms CM3-large, which has over 30x more parameters, for half of the datasets. BART+KILM$_{\text{DU}}$ outperforms CM3-large in four out of six datasets. 
CM3 as a PLM has an impressive performance on entity disambiguation task with no additional training, and this comparison shows that BART+KILM can outperform CM3 with much less parameters. We also present results comparing BART+KILM with BLINK~\citep{wu2020scalable} in \Cref{tab:el2}, where we see that it performs competitively compared to BLINK (which is fine-tuned for entity disambiguation). Moreover, the large gap between the performance of BART+KILM and BART+Merge shows that the proposed distinct structure (and not necessarily the data) plays a key role in the performance of BART+KILM in this task.

\begin{table}[t!]
\centering
\resizebox{.9\linewidth}{!}{
\begin{tabular}{@{}lcccccc@{}}
\toprule
\multirow{2}{*}{\textbf{Model}} & \multicolumn{3}{c}{\textbf{News ORG}} & \multicolumn{3}{c}{\textbf{News PER}} \\ \cmidrule(l){2-4} \cmidrule(l){5-7} 
 & Ap. & Pref. & NH. & Ap. & Pref. & NH. \\ \midrule
BART-base & 26.0 & 17.8 & 41.7 & 48.0 & 14.3 & 28.3 \\
\quad+KILM & \textbf{97.0} & \textbf{51.5} & \textbf{56.8} & \textbf{94.0} & \textbf{36.0} & \textbf{42.0} \\ \midrule
\multirow{2}{*}{\textbf{Model}} & \multicolumn{3}{c}{\textbf{Wiki ORG}} & \multicolumn{3}{c}{\textbf{Wiki PER}} \\ \cmidrule(l){2-4} \cmidrule(l){5-7} 
 & Ap. & Pref. & NH. & Ap. & Pref. & NH. \\ \midrule
BART-base & 48.5 & 26.7 & 49.7 & 30.8 & 7.3 & 32.7 \\
\quad+KILM & \textbf{98.0} & \textbf{48.0} & \textbf{61.0} & \textbf{89.9} & \textbf{40.3} & \textbf{50.3} \\ \bottomrule
\end{tabular}
}
\caption{Human evaluation results on Appositive Generation in News and Wikipedia domains on org- and person-type entities (see \Cref{sec:appx:human}). \textit{Ap.}, \textit{Pref.}, and \textit{NH.} mean \textit{Is Appositive}, \textit{Preference}, and \textit{Not Hallucinated}. Numbers in bold are significantly better than those from BART at p-value of 0.05 in a pairwise t-test.}
\label{tab:human:appos}
\end{table}

\paragraph{Appositive Generation}
\label{sec:appos}
Appositive generation is the task of adding background information for named entities in a sentence in the form of an appositive phrase. As shown in \Cref{tab:examples}, we construct structured prompts to probe PLMs without fine-tuning on ApposCorpus~\citep{kementchedjhieva2020apposcorpus}. We consider the generated texts recovered from the mask tokens in the short description field as the generated appositives.\footnote{Since the pre-training corpus of BART includes Wikipedia articles, BART can also recover appositives from mask tokens without further task adaptation.}

Since automatic metrics only assess the text overlap based performance (\Cref{tab:appos} in \Cref{sec:appx:appos} with comparisons with SOTA), we conduct human evaluation for a more comprehensive evaluation from three aspects: \textit{Is Appositive} (Ap.), \textit{Preference} (Pref.), and \textit{Not Hallucinated} (NH.). \textit{Ap.} evaluates whether the generation is an appositive or not, while \textit{Pref.} evaluates the suitability of the generated appositives to the context. \textit{NH.} evaluates whether the model generates a hallucinated appositive or not, verifying whether the generated appositive is factually correct.
Pairwise A/B testing is utilized to compare the performances of BART before and after KILM (in the primary setting) on all four subsets of ApposCorpus. 
For each comparison, the same context and two options generated by models for comparison are first randomly shuffled and then are shown to the annotators. Each comparison requires three judgments. 50 data samples are randomly selected from each subset.
More details of human evaluation are included in \Cref{sec:appx:human}. 
\Cref{tab:human:appos} lists the human evaluation results in terms of the winning rate (ties are counted as wins for both), where we observe that BART+\kilm~generates better appositives and hallucinates less in all four subsets. These results indicate that BART+\kilm~possesses more entity-related knowledge than BART.

\paragraph{In-Context Few-Shot QA}

The implicit knowledge embedded in the parameters can support large PLMs to obtain competitive results on open-domain QA tasks without accessing external knowledge~\citep{roberts2020much,radford2019language,brown2020language}. 
We conduct in-context few-shot experiments, in the primary setting of KILM, on TriviaQA~\citep{joshi2017triviaqa}, Natural Questions (NQ)~\citep{kwiatkowski2019natural}, and Web Questions (WQ)~\citep{berant2013semantic} datasets. Similar to the settings of GPT-3~\citep{brown2020language}, we put several example QA pairs into the input sequences of both the encoder and decoder. The format of prompting is shown in \Cref{tab:examples}, while the example QA pairs are retrieved with a TF-IDF retriever\footnote{The implementation is based on \url{https://github.com/efficientqa/retrieval-based-baselines}.} from the corresponding training set. The tokens recovered from the mask tokens from the decoder will be considered as the generated answers.


\begin{figure}[t!]
  \centering
  \includegraphics[width=\linewidth]{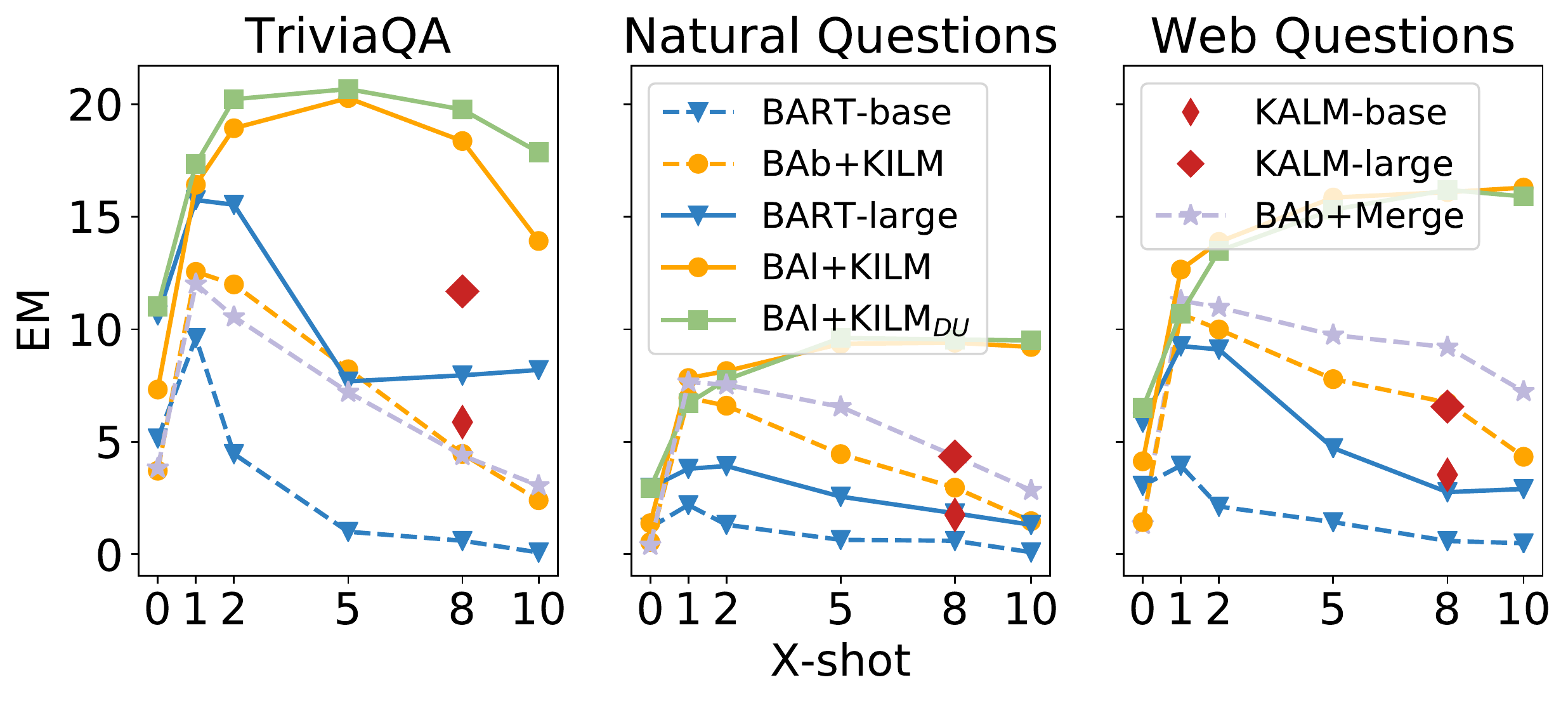} 
  \caption{Results on QA datasets with different shots. BART results are in blue, while the results of BART+KILM are in orange and green. We use dashed and solid lines to denote the base- and large-size models, respectively. Also ``BAb'' and ``BAl'' correspond to BART-base and BART-large, respectively. KILM$_{DU}$ is KILM with data upscaling where entire Wikipedia articles are used instead of only their first paragraphs.}
  \label{fig:qa}
\end{figure}
We illustrate learning trends with different ``shots'' in \Cref{fig:qa} on all three datasets. Interestingly, BART+KILM mostly performs worse than the original BART under the zero-shot setting. However, appending demonstrations into the contexts enables BART+KILM to outperform the original BART by a large margin. With the data upscaling setting, KILM$_{\text{DU}}$ provides comparable (or even larger) improvements to BART under the few-shot setting while slightly improving the zero-shot performances of BART. 
Though far from perfect, these results suggest that KILM significantly improves the in-context learning ability of BART on all three QA datasets. KILM also enables BART to pack factoid knowledge more effectively within its parameters, which supports QA. BART-base+KILM outperforms BART-large under the in-context few-shot setting for the NQ and WQ datasets. The performance of the baseline model, BART+Merge, shows a similar trend to BART+KILM with little advantage on NQ and WQ datasets. This indicates that pre-training with data in ``<Entity> is <Short Desc>'' format is more suitable for QA tasks. Nevertheless, the proposed distinct structure does not bring much obstacle to BART+KILM on QA tasks.

\begin{table}[t!]
    \centering
    \resizebox{\linewidth}{!}{
    \begin{tabular}[t]{@{}lcccccc@{}}
    \toprule
    \multirow{2}{*}{Model} & \multicolumn{3}{c}{Seen Test} & \multicolumn{3}{c}{Unseen Test} \\ \cmidrule(l){2-4} \cmidrule(l){5-7} 
        & PPL    &    R1      & R2      & PPL     & R1      & R2      \\ \midrule
    SKT & 52.0 & 19.3 & 6.8 & 81.4 & 16.1 & 4.2 \\
    KAT-TSLF & 14.4 & 21.7 & 7.6 & 15.8 & 20.7 & 7.2 \\ \midrule
    BART-base             & \textbf{17.1}$^*$     & 18.7          & 4.9           & \textbf{20.9}$^*$       & 17.5     & 4.0           \\
    \quad+Merge           & 21.4      & \textbf{19.3}       & \textbf{5.2}           & 26.8         & \textbf{18.0}     & \textbf{4.2}           \\
    \quad+KILM            & 21.5      & \textbf{19.3}$^*$          & \textbf{5.2}           & 26.9         & 17.9$^*$     & \textbf{4.2}$^*$           \\\midrule
    BART-large            & \textbf{14.2}$^*$      & 20.6          & 5.8           & \textbf{18.7}       & 18.5           & 4.3           \\
    \quad+KILM            & 18.9      & \textbf{20.8}$^*$          & \textbf{5.9}           & 24.9         & \textbf{18.8}$^*$           & \textbf{4.5}$^*$           \\ \bottomrule
    \end{tabular}
    }
    \caption{WoW test set results. \textit{PPL} denotes perplexity, while \textit{R1/2} denotes ROUGE-1/2 metrics. While both SKT~\citep{kim2019sequential} and KAT-TSLF~\citep{liu2021three} use external knowledge as inputs, BART and BART+KILM are evaluated without knowledge to better demonstrate the impact of KILM. $^{*}p<0.05$ in a pairwise t-test for comparison between ours and BART.}
    \label{tab:wow}
\end{table}
\begin{table}[t!]
    \centering
    \resizebox{.95\linewidth}{!}{
    \begin{tabular}[t]{@{}lcccccc@{}}
    \toprule
    \multirow{2}{*}{Model} & \multicolumn{3}{c}{Seen Test} & \multicolumn{3}{c}{Unseen Test} \\ \cmidrule(l){2-7} 
                           & Flu.  & Info.  & NH. & Flu.   & Info.  & NH.  \\ \midrule
    BART-base             & 59.7     & \textbf{64.0}   & 48.4        & 65.8      & \textbf{70.3}   & 46.6         \\
    \quad+KILM              & \textbf{66.7}     & 63.0   & \textbf{60.3}$^*$        &  \textbf{69.2}       & 69.3   & \textbf{58.8}$^{\dagger}$  \\ \bottomrule
    \end{tabular}
    }
    \caption{Human evaluation results on WoW test sets without external knowledge inputs. Flu., Info., and NH. are \textit{Fluency}, \textit{Informativeness} and \textit{Not Hallucinated} respectively. $^{*}$Model performs significantly better than the baseline ($p < 0.05$); $^{\dagger}$Pairwise t-test ($p < 0.07$).}
    \label{tab:human:wow}
\end{table}

\begin{table}[t]
\centering
\resizebox{.7\linewidth}{!}{%
\begin{tabular}{@{}lccc@{}}
\toprule
\multirow{2}{*}{Model}    & \textbf{GLUE} & \textbf{CNN}     & \textbf{XSUM}              \\
                          & Avg.  & R1 & R1      \\ \midrule
BART-base                 & 83.3 & 42.79 & \textbf{40.83}$^*$ \\
\quad+KILM                & \textbf{83.8} & \textbf{42.86} & 40.76 \\ \midrule
BART-large                & 87.1 & \textbf{44.14}$^*$ & \textbf{45.17} \\
\quad+KILM                & \textbf{87.7} & 43.15 & 45.07 \\ \bottomrule
\end{tabular}
}
\caption{Results on the GLUE and summarization test sets. We report average score of Matthews correlation for CoLA and accuracy scores for other tasks in GLUE benchmark; and ROUGE-1 for summarization. 
$^{*}$pairwise t-test $p < 0.05$.}
\label{tab:glue+summ:simple}
\end{table}

\paragraph{Knowledge Grounded Response Generation (KGRG)} 
The KGRG task requires topical and factual knowledge~\citep{petroni2021kilt} for a chatbot to make engaging conversations with users on various topics~\citep{ghazvininejad2018knowledge}. 
We fine-tune BART before and after KILM on the Wizard of Wikipedia (WoW)~\citep{dinan2018wizard} dataset without using knowledge as input, to better study the impact of the injected knowledge under a knowledge-unavailable setting. 
The generated responses are evaluated with PPL, ROUGE-1 and ROUGE-2 metrics. In \Cref{tab:wow}, BART+KILM offers a consistent and significant advantage over BART on ROUGE scores,
whereas it underperforms BART on PPL. The performance gap on PPL can be attributed to 
the fact that many of the responses in WoW contain hallucination ~\citep{dziri2022origin}, which is somewhat mitigated by KILM.
Compared to the strong baseline with external knowledge inputs, BART+KILM even performs comparably with SKT~\citep{kim2019sequential}. Note that the performance of BART+Merge shows no  difference from BART+KILM, which suggests that the introduced distinct structure does not affect BART's application of injected knowledge on WoW.

While automatic metrics are important in KGRG evaluation, they do not always tell the whole story~\cite{DBLP:journals/corr/abs-2106-06411}, therefore we also conduct human evaluation on WoW test sets  from three aspects, namely \textit{Fluency (Flu.)}, \textit{Informativeness (Info.)}, and \textit{Not Hallucinated (NH.)}. \textit{Flu.} focuses on whether the responses are fluent and consistent with respect to the conversation so far, while \textit{Info.} evaluates whether the responses contain verifiable factual information. The evaluation on \textit{NH.} is only valid when a response is informative. The settings of human evaluation are the same as those for appositive generation
(see \Cref{sec:appx:human}). The results in \Cref{tab:human:wow} demonstrate that BART+KILM performs comparably with BART in terms of fluency and informativeness, while it tends to \textbf{hallucinate less} when generating factual information in the responses, especially in unseen domains.


\subsection{General Tasks}
\label{sec:gen_tasks}

We now evaluate the impact of \kilm~on models' performance on general NLU and NLG tasks using the GLUE benchmark~\cite{wang2018glue} and summarization datasets, CNN/Dailymail~\citep{hermann2015teaching} and XSUM~\citep{narayan2018don}, by fine-tuning both BART and BART+KILM for comparison. The summary of the results is shown in~\Cref{tab:glue+summ:simple}, and the detailed results shown in \Cref{tab:glue} and \Cref{tab:summ_full}. 
BART+KILM outperforms BART marginally on GLUE and the differences for summarization datasets are small.
These results suggest that \kilm~preserves the performance of the original BART on downstream NLU and NLG tasks, and even in some cases it improves it. They also verify that KILM does not cause catastrophic forgetting of the original learnings in BART, thus making BART+KILM a reliable PLM.

\section{Discussions}
\label{sec:discussion}
\paragraph{Roles of Introduced Special Tokens}
The introduced special tokens to mark beginning and end of entities (\texttt{<ent>}, \texttt{</ent>}) and entity descriptions (\texttt{<ent\_desc>}, \texttt{</ent\_desc>}) form a distinct structure in pre-training samples, which inserts entity-centric knowledge into pre-training corpora, thus injects knowledge in PLMs. We discuss the roles of these special tokens from the following aspects:

\textbf{\textit{Entity Knowledge Probing:}}
This distinct structure in KILM provides a tool for probing the entity-related knowledge retained in PLMs. To demonstrate this, we probe BART+KILM by prompting it to generate short descriptions for entities in validation set\footnote{The articles in validation set are not included in the pre-training process, whereas the involved entities mostly are.} of the pre-training corpus.
The probing format and the corresponding results are shown in \Cref{sec:short_desc} and \Cref{tab:desc_gen}. BART+KILM achieve around 60 unigram F1 scores with no performance gap with the data samples from a subset of the training set. These results indicate that we can easily recall the entity description knowledge in different contexts without sensitivity to prompt designs.
It is shown that the proposed pre-training structure is the \textbf{main contributor} of the improvements on entity-related datasets, especially in zero-shot manner. By leveraging the introduced special tokens, the knowledge retained in PLMs can be more efficiently leveraged on downstream tasks.

\textbf{\textit{Structured Prompt:}}
The special tokens also provide convenient knowledge probing for zero-shot entity-centric tasks, such as entity disambiguation and appositive generation (\Cref{sec:ed}). 

For additional discussions on the need for special tokens, please refer to \Cref{sec:appx:discussion}.

\paragraph{Is KILM's impact equal on different domains and tasks?}
\label{sec:exp:analysis}
Despite the above-mentioned gains, BART+\kilm~appears to be less knowledgeable than BART on person-type entities, as manifested in the performance gap between organization- and person-type entities in appositive generation (\Cref{tab:human:appos}).
That may be due to the type of knowledge content injected by KILM. 
The entity knowledge required for generating appositives varies vastly from biographies to relationships with other people.
However, short descriptions in Wikipedia for person-type entities focus mostly on their nationality and occupation. Also, many of them are similar~\footnote{For example short descriptions for both \textit{Columbus Short} and \textit{Drew Fuller} are \textit{``American actor''}}. This problem also affects the performance in \Cref{tab:lama} on G-RE datasets in LAMA benchmark. More analyses are in \Cref{sec:lama}. We leave the study of enriching the knowledge content for pre-training as future work.

The proposed pre-training structure shows its strength in entity-related tasks. Nevertheless, 
KILM may downgrade to conventional knowledge-augmented pre-training (BART+Merge) when the pre-training objective of KILM is not fully aligned with those of the downstream tasks.
 
\paragraph{Placement of Knowledge Component}
An ablation study on the knowledge component placement in KILM is presented in \Cref{sec:ablation}, where we show that putting short descriptions right after entity mentions results in better performance compared to placing them at the end of sentences.

\paragraph{Extending KILM for Other PLM Architectures}
\label{sec:other-plm}
In this paper, we choose BART as the default PLM; however, KILM can also be applied to other PLMs by adjusting their training objectives for knowledge infilling. For decoder-only PLMs, such as GPT-2, the knowledge component, i.e., short descriptions,
can be moved to the end of the target sequence (similar to CM3) instead of being adjoined the surface form of the entity. As for encoder-only PLMs, such as BERT, contrastive training strategy introduced in LinkBERT~\citep{yasunaga2022linkbert} is one option for the training objective of KILM. 
Due to the substantial computational cost of training these models, we leave these explorations for future works.

\section{Conclusion}

In this paper, we propose a novel method, KILM, to inject entity-related knowledge into large PLMs through \continuedtraining. Our approach enhances the performance of the original PLMs on knowledge-intensive tasks, especially in zero- and few-shot settings, while not causing catastrophic forgetting of the knowledge in the origianl PLMs. 

\clearpage

\bibliography{KILM}
\bibstyle{acl_natbib}

\appendix

\section{Limitations}
In this paper, we propose a continued pre-training method to inject knowledge into large-pre-trained language models. There are eight V100 GPUs involved in each pre-training experiments and the whole pre-training process takes 5 days for the base-size model and 13 days for the large-size model, in primary settings. These numbers in data upscaling settings are significantly greater (30 days for the large-size model). Despite its advantage in reducing resource need in inference time, KILM is both time-consuming and computational resource-consuming during the training time. 

Similar to any model-based generation systems, KILM could be prone to generating factually incorrect statements with regards to entities. These statements might also be prone to be biased based on ethnicity, race , and sexual orientation.

\section{Additional Discussions}
\label{sec:appx:discussion}
\paragraph{Are New Special Tokens Needed?}
There are a few reasons for introducing new special tokens in KILM for marking entities and their descriptions instead of reusing existing tokens, such as commas or parentheses. 
First, many entities have commas and parentheses in their names, making the entity descriptions indistinguishable from the contexts. For instance, there are 378,093 entities in English Wikipedia with a comma in their names, such as the entity \textit{``Mars, Aurgazinsky District, Republic of Bashkortostan''}. 
Second, using commas or parentheses could break the fluency of the text. In a context like ``\textit{The Baltic states [...] is used to group three countries: Estonia, Latvia, and Lithuania}'', adding a short description for the entity ``Estonia'' using a comma would break the fluency of the sentence. 
Finally, using commas or parenthesis will overload their meanings, and during prompting of the model for knowledge probing it will result in a lack of clarity for the model as to how the comma or parenthesis should be interpreted. 

\paragraph{Justifications on the additional cost during pre-training}
Injecting additional knowledge text into pre-training corpora may introduce additional costs during the pre-training process. While entity descriptions used in the paper are usually a one-sentence definition of an entity, the average length of short descriptions is 13.81 words. Considering that we split the Wikipedia articles with document strides of 512, the inserted tokens for short descriptions only take 2.6\% of the length of the whole sequence, which does not bring much more training cost.

\section{Analysis}
\subsection{Entity Description Probing}
\label{sec:short_desc}

We analyze the quality of the knowledge injection process by evaluating the model's performance on entity description probing with structured prompts. This task is aligned with our proposed pre-training objective and reflects the effect of the continued pre-training. This can be considered as a plug-and-play process for knowledge induction by simply inserting the proposed distinct structure. We conduct evaluation on the validation set and a subset of the training set with around 10k data samples of our pre-training corpus. The data samples in the training subset are randomly selected, whereas the data samples in the validation set are not included in the training process. More specifically, the entities in the validation set may appear in the training set. However, the contexts of the entities in the paragraphs do not.
We demonstrate the structured prompts for entity description probing as follows:
\begin{quote}
\textbf{Input/Prompt:} \texttt{\underline{The Joker is a} \underline{comic book series published by DC} \underline{Comics starring the supervillain} \underline{the}} \colorbox{myorange}{\underline{\texttt{<ent>}}} \underline{Joker} \texttt{\colorbox{myorange}{\underline{</ent><ent\_desc>}}} \colorbox{myblue}{<mask>} \colorbox{myorange}{</ent\_desc>}. \\
\textbf{Target:} \texttt{Joker (character) <sep> Fictional character throughout the DC Universe} 
\end{quote}

The example illustrates the input sequence of the encoder, while the prompt to the decoder is the same until the \texttt{<ent\_desc>} token (marked with \underline{underline}). Similar to the decoder-only models, the model is expected to continue generating entity descriptions following the prompt, until the \texttt{</ent\_desc>} token is generated.

The generated entity descriptions are evaluated with exact match (EM) and unigram F1 scores. As the results are shown in \Cref{tab:desc_gen}, for KILM in the primary setting, BART models with KILM achieve around 40 EM and 60 F1 scores. Interestingly, there is a marginal performance gap between the seen and unseen validation sets. The results indicate our model not only embed the knowledge with its parameters, but also can recall the injected knowledge under unseen contexts without much performance loss. 

\setcounter{table}{0}
\renewcommand{\thetable}{C\arabic{table}}
\begin{table}[t!]
\centering
\resizebox{\linewidth}{!}{
\begin{tabular}{@{}lcccc@{}}
\toprule
\multirow{2}{*}{\textbf{Model}} & \multicolumn{2}{c}{\textbf{Train subset}} & \multicolumn{2}{c}{\textbf{Valid}} \\ \cmidrule(l){2-5} 
                      & EM                 & F1                 & EM                  & F1                  \\ \midrule
BART-base + \kilm              & 37.75              & 58.08              & 37.60               & 58.48               \\
BART-large + \kilm             & 42.58              & 61.96              & 42.84               & 62.69               \\ \midrule
BART-large + \kilm$_{\text{End}}^\dagger$ & 38.64 & 57.97 & 38.59 & 57.71 \\
\bottomrule
\end{tabular}
}
\caption{Results of short description generation on a subset of the training set and the validation set of the pre-training corpus. $^\dagger$\kilm$_{\text{End}}$ is a variant of KILM for ablation study (\Cref{sec:ablation}).}
\label{tab:desc_gen}
\end{table}

\begin{table}[t!]
\centering
\resizebox{.95\linewidth}{!}{
\begin{tabular}[t]{@{}lcccc@{}}
\toprule
\multirow{2}{*}{\textbf{Model}} & \multirow{2}{*}{\textbf{G-RE}} & \multirow{2}{*}{\textbf{T-REx}} & \multirow{2}{*}{\textbf{C-Net}} & \multirow{2}{*}{\textbf{SQuAD}} \\ 
&&&&\\\midrule
BERT-base & 9.12 & 30.83 & 14.29 & 15.88 \\
ERNIE     & 6.62 & 27.58 & 13.62 & 14.83 \\
LM-CORE     & \underline{23.13} & \underline{55.32} & \underline{17.28} & \underline{16.15} \\ 
KALM-base   & 3.27 & 25.96 & 8.61 & 6.64 \\
KALM-large  & 5.41 & 28.12 & 10.70 & 11.89 \\ \midrule
BART-base & \textbf{5.70} & 22.14 & \textbf{13.88} & 6.29 \\
\quad+Merge & 5.50 & \textbf{24.98} & 13.03 & 7.69 \\ 
\quad+KILM & 4.02 & 23.41 & 12.80 & \textbf{8.39} \\ \midrule
BART-large & \textbf{7.76} & 26.00 & 16.07 & 11.19 \\
\quad+KILM & 6.83 & \textbf{26.14} & \textbf{16.96} & 11.19 \\ 
\quad+KILM$_{\text{DU}}$ & 3.10 & 24.99 & 16.22 & \textbf{12.94} \\ 
\bottomrule
\end{tabular}
}
\caption{Accuracy on the LAMA benchmark. The best results are marked with \underline{underline}, while \textbf{Bold} indicates the better result of comparison between BART before and after KILM. The results of previous models except BART are taken from \cite{zhang2019ernie,rosset2020knowledge,kaur2022lm}.}
\label{tab:lama}
\end{table}

\subsection{LAMA Knowledge Probing}
\label{sec:lama}

\citet{petroni2019language} proposed the LAMA benchmark to provide an in-depth study of relational knowledge in PLMs by probing the answers to ``fill-in-the-blank'' cloze statements. Different types of relational knowledge are evaluated with statements semi-manually constructed from different knowledge sources, including Google-RE (G-RE), T-REx~\citep{elsahar2018t}, ConceptNet (C-Net)~\citep{speer2012representing} and SQuAD~\citep{rajpurkar2016squad}. We follow the original LAMA settings, while only keeping the data samples whose answer length is 1 after tokenization. The probing input and output format of BART and BART+KILM is shown as followings:

\begin{quote}
\textbf{Input/Prompt:} \texttt{\underline{The Teatr Wielki }} \texttt{\underline{is a} \colorbox{myblue}{<MASK>}.} \\
\textbf{Target:} \texttt{theatre}
\end{quote}

Similar to entity description probing in \Cref{sec:short_desc}, \textit{``Input''} and \textit{``Prompt''} (with underline) are inputs to BART encoder and decoder, respectively. The generation is considered to be correct only if it is exactly the same with \textit{``Target''}.
We present the probing results in \Cref{tab:lama}. We also include the results of BERT~\citep{devlin2019bert}, BERT-based ERNIE~\citep{zhang2019ernie}, BERT-based LM-CORE~\citep{kaur2022lm}, and GPT-2-style KALM~\citep{rosset2020knowledge} for reference. However, because of the differences on the tokenization and pre-training process, different PLMs are not comparable on LAMA benchmark~\citep{jiang2020can}. Even though KILM does not inject relational knowledge into PLMs, we still observe improvements after KILM on all the datasets except G-RE. As it's discussed in \Cref{sec:exp:analysis}, the injected knowledge of person-type entities is not aligned with the knowledge required by G-RE, since the samples from G-RE are focused on \texttt{date\_of\_birth} and \texttt{place\_of\_birth} relations in the \textit{person} domain. Under the data upscaling setting, KILM$_{\text{DU}}$ further enhances the rational knowledge required for SQuAD, while LAMA performance is negatively impacted for other datasets.
The results indicate that injecting the entity description knowledge also helps models better understand the relationships between specific entities. Moreover, the results of KILM$_{\text{DU}}$ suggest that the injected knowledge has closer relevance to the knowledge for SQuAD, whereas far from that of G-RE and T-REx. 

\setcounter{figure}{0}
\renewcommand{\thefigure}{C\arabic{figure}}
\begin{figure*}[t]
  \centering
  \includegraphics[width=\linewidth]{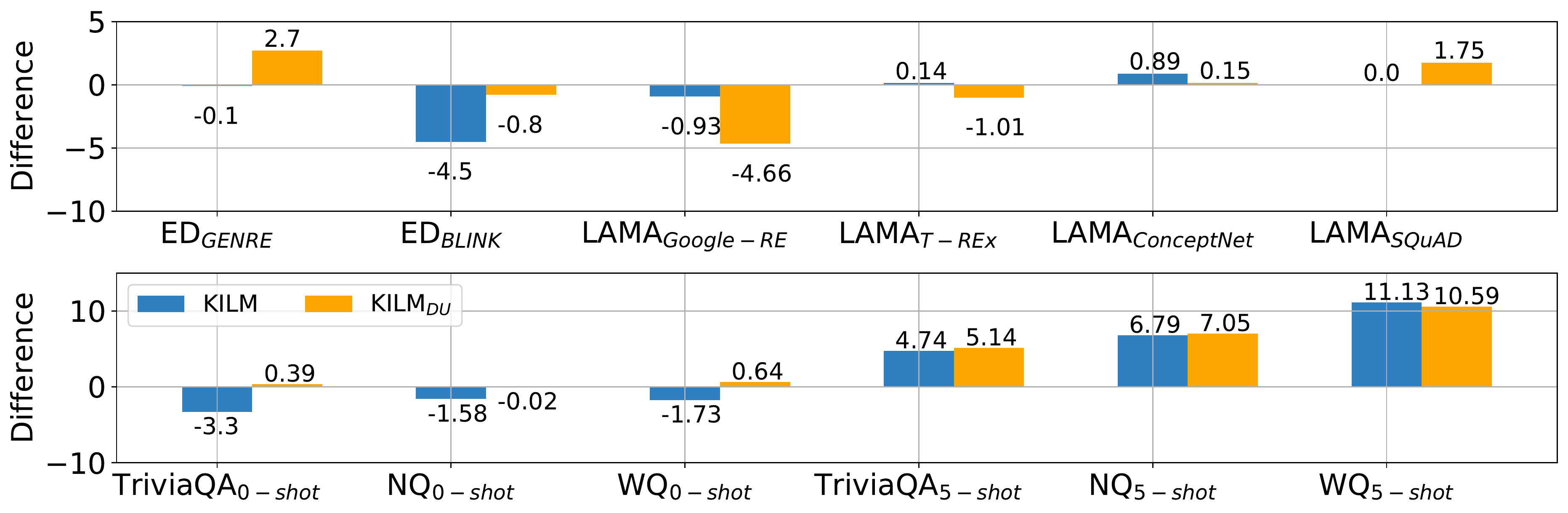} 
  \caption{The performance difference between BART-large+KILM (or KILM$_{\text{DU}}$) and the corresponding baseline models on entity disambiguation, LAMA and QA (TriviaQA, NQ, and WB) tasks. More specifically, the baseline models of entity disambiguation tasks are CM3-large and BLINK with GENRE and BLINK candidates, while the baseline model of both LAMA and QA tasks is the original BART-large. We also display the performance differences along with each bar, where a positive number denotes a better performance of BART+KILM vs the baseline.}
  \label{fig:data_scaling}
\end{figure*}

\subsection{Ablation Study}
\label{sec:ablation}

We conduct an ablation study on the knowledge component position in KILM. We compare our method with KILM variant that moves the knowledge component (highlighted in blue in \Cref{fig:object}) including \texttt{<ent\_desc>} and \texttt{</ent\_desc>} to the end of the target sequence. The variant of the target sequence in \Cref{fig:object} is as follows:

\begin{quote}
    \texttt{The Joker is a comic book series published by DC Comics starring the supervillain the \colorbox{myorange}{<ent>} Joker \colorbox{myorange}{</ent>}. It ran for nine ... </s></s>\colorbox{myorange}{<ent\_desc>} \colorbox{myblue}{Joker (character)<sep>Fictional} \colorbox{myblue}{character throughout the DC} \colorbox{myblue}{Universe} \colorbox{myorange}{</ent\_desc>}}
\end{quote}

We denote this KILM variant as KILM$_{\text{End}}$. We evaluate these two models on entity description probing and zero-shot entity disambiguation tasks. As shown in \Cref{tab:desc_gen} and \Cref{tab:el2}, BART with KILM consistently outperforms BART with KILM$_{\text{End}}$ on both tasks. Despite the performance gap, the advantage of KILM$_{\text{End}}$ is that KILM$_{\text{End}}$ can also be applied to decoder-only models, such as GPT-2, for entity knowledge injection.

\subsection{Data Scaling Laws}

As mentioned in \Cref{sec:data}, we conduct continued pre-training under two settings: the primary setting and the data upscaling setting. While the primary setting only uses the paragraphs in Wikipedia summary sections, the data upscaling setting extends the training corpus to the whole Wikipedia corpus, which enlarges the training set by more than two million data samples and double the pre-training time. To study the effect of data scaling, we compare the performances of BART-large+KILM under primary and data upscaling settings on knowledge-intensive tasks, including entity disambiguation, LAMA, and closed-book QA tasks. The evaluation on entity disambiguation tasks involves six datasets and we only compare the average InKB F1 scores, since during data scaling, the performances are consistently improved across all the datasets. 

In \Cref{fig:data_scaling}, we show the performance difference between BART-large+KILM ( or KILM$_{\text{DU}}$) and the corresponding baseline models on entity disambiguation, LAMA (in the first row) and QA (including three datasets under 0/5-shot in the second row) tasks. We also display the performance differences along with each bar, where a positive number denote a better performance of BART+KILM. According to the comparison, KILM in both settings shows little benefit for Google-RE and T-REx datasets in LAMA benchmark and makes it harder for the model to recall the relational knowledge in specific domains. On the other hand, for the entitiy-based tasks, such as entity disambiguation, the injected knowledge through KILM equip BART with great zero-shot ability, comparing to the strong baseline models, which we've discussed in \Cref{sec:ed}. For QA tasks, BART+KILM in the primary setting performs worse than the original BART model in a zero-shot manner, however, BART+KILM in data upscaling setting works comparably with the original BART in this case. Together all these comparisons, we conclude that KILM, as a proposed novel technique for entity-related knowledge injection, is able to largely benefit the model in terms of zero-shot ability on entity-based knowledge-intensive tasks. However, even though we jointly pre-train the model with the original text infilling objective of BART, catastrophic forgetting of some specific knowledge is unavoidable, especially in the data upscaling setting.

\subsection{Case Study}

Some selected data sample from ApposCorpus and WoW are shown in~\Cref{tab:example:appos} and~\Cref{tab:example:wow}. For zero-shot appositive generation task, while the original BART-base model tends to generate appositives with similar surface forms to the gold ones or a piece of text that fit the context, it hallucinates a lot.
BART-base+KILM is more knowledgeable on the actual meaning of the entities, however, it still make mistakes in terms of the date and specific occupation. For KGRG task with task-specific training, both models are able to generate fluent responses. At the same time, BART+KILM tends to hallucinate less by including a bit less information in some cases.

\section{KILM Algorithm}
\label{sec:appx:alg}

We denote the data transformations of the text infilling and sentence permutation objectives for BART as \textsc{TextMask} and \textsc{SentPerm}.
In the original pre-training process of BART, given a target sequence with $M$ tokens $\mathbf{Y} = \{t_1, t_2, ..., t_M\}$, and the corresponding corrupted input sequence $\mathbf{X} = \{t'_1, t'_2, ..., t'_N\}$ with $N$ tokens, the model, parameterized by $\theta$, is optimized by minimizing the reconstruction loss over the whole sequence $\mathbf{Y}$:

\setlength{\abovedisplayskip}{1pt}
\setlength{\belowdisplayskip}{1pt}

\begin{align}
    \mathbf{X} &= \textsc{SentPerm}(\textsc{TextMask}(\mathbf{Y})) \\
    \mathcal{L} &= \mathbb{E}(\sum_{m=1}^{M}-\log  p(t_m|t_{1:m-1}, \mathbf{X}, \theta)). 
    \label{eq:general}
\end{align}

For the proposed KILM \continuedtraining, the original document, the selected entity, and the corresponding injected knowledge are represented as $\mathbf{S}=\{t_1, t_2, ..., t_N\}$, $E$, and $\mathbf{K}=\{k_1, k_2, ..., k_L\}$, respectively. The data transformation procedure can be represented as 
\begin{align}
    \mathbf{Y} &= \textsc{KnInfill}(\mathbf{S}, E, \mathbf{K}), \label{eq:entity_transfer1} \\
    \mathbf{X} &= \textsc{KnMask}(\mathbf{Y}). \label{eq:entity_transfer2}
\end{align}
The final loss can be denoted as:

\setlength{\abovedisplayskip}{2pt}
\setlength{\belowdisplayskip}{2pt}

\begin{equation}
    \mathcal{L} = (1-\alpha-\beta)\mathcal{L}_{copy} + \alpha\mathcal{L}_{infill} + \beta\mathcal{L}_{kn},
    \label{eq:KILM}
\end{equation}
where $\alpha$ and $\beta$ are calculated based on the proportion of the corresponding spans across the entire sequence.
The resulting KILM algorithm for continual pre-training is summarized in Algorithm~\ref{alg:general}.

\begin{algorithm}[t]
\caption{KILM Pre-training Process}\label{alg:general}
\KwIn{Model $\mathrm{M_{\theta}}$, Number of Epochs $\mathit{T}$, Wikipedia Corpus $\mathbb{S}$, Knowledge Corpus $\mathbb{K}$.}
\For{$i=1$ \KwTo $\mathit{T}$}
{   
\For {each $S_j \in \mathbb{S} $}
{
Sample one entity $E_j^i$ from $S_j$; \\ 
Retrieve entity knowledge: \\
$\mathbf{K} = \textsc{LookUp}(\mathbb{K}, E_j^i)$; \\
Construct training samples: \\
$\mathbf{Y_j^i} = \textsc{KnInfill}(\mathbf{S_j}, E_j^i, \mathbf{K})$, \\
$\mathbf{X_j^i} = \textsc{TextMask}(\textsc{KnMask}(\mathbf{Y_j^i}))$; \\
Optimize $\mathrm{M_{\theta}}$ with Eq.~\ref{eq:KILM}.
}
}
\end{algorithm}

\setcounter{table}{0}
\renewcommand{\thetable}{E\arabic{table}}

\begin{table*}[t!]
\centering
\resizebox{.93\linewidth}{!}{%
\begin{tabular}{@{}lcccccccr@{}}
\toprule
\textbf{Models} & \textbf{AIDA} & \textbf{MSNBC} & \textbf{AQUAINT} & \textbf{ACE2004} & \textbf{CWEB} & \textbf{WIKI} & \textbf{Avg} & \textbf{\#Params} \\ \midrule
BLINK$^\dagger$ & 79.6 & \textbf{80.0} & \textbf{80.3} & 82.5 & \textbf{64.2} & \textbf{75.5} & \textbf{77.0} & 336M \\
BART-base      & 18.3 & 30.8 & 8.7 & 20.3 & 23.7 & 20.5 & 20.4 & 139M  \\
BART-base+Merge  & 19.5 & 24.1 & 12.2 & 18.4 & 21.9 & 19.8 & 19.3 & 139M \\
BART-base+KILM & 75.1 & 69.3 & 67.8 & 77.4 & 57.4 & 62.2 & 68.2 & 139M \\
BART-large      & 17.4 & 39.1 &  9.6 & 27.4 & 26.6 & 21.5 & 23.6 & 406M  \\
BART-large+KILM & 80.1 & 75.2 & 71.0 & 82.4 & 60.0 & 66.5 & 72.5 & 406M \\
BART-large+KILM$_{\text{DU}}$   & \textbf{82.1} & 76.4 & 77.8 & \textbf{86.4} & 62.4 & 72.3 & 76.2 & 406M \\ 
\midrule
BART-large+KILM$_{\text{End}}^\ddagger$  & 79.6 & 74.5 & 69.6 & 82.1 & 59.2 & 64.2 & 71.5 & 406M \\
\bottomrule
\end{tabular}
}
\caption{InKB Micro F1 on zero-shot entity disambiguation tasks with BLINK candidates. $^\dagger$The results are taken from \url{https://github.com/facebookresearch/BLINK} and normalized over the whole dataset. $^\ddagger$\kilm$_{\text{End}}$ is a variant of KILM for ablation study (\Cref{sec:ablation}). \textbf{\#Params} denotes the number of parameters of the models.} 
\label{tab:el2}
\end{table*}

\setcounter{figure}{0}
\renewcommand{\thefigure}{E\arabic{figure}}
\begin{figure}[t!]
  \centering
  \includegraphics[width=\linewidth]{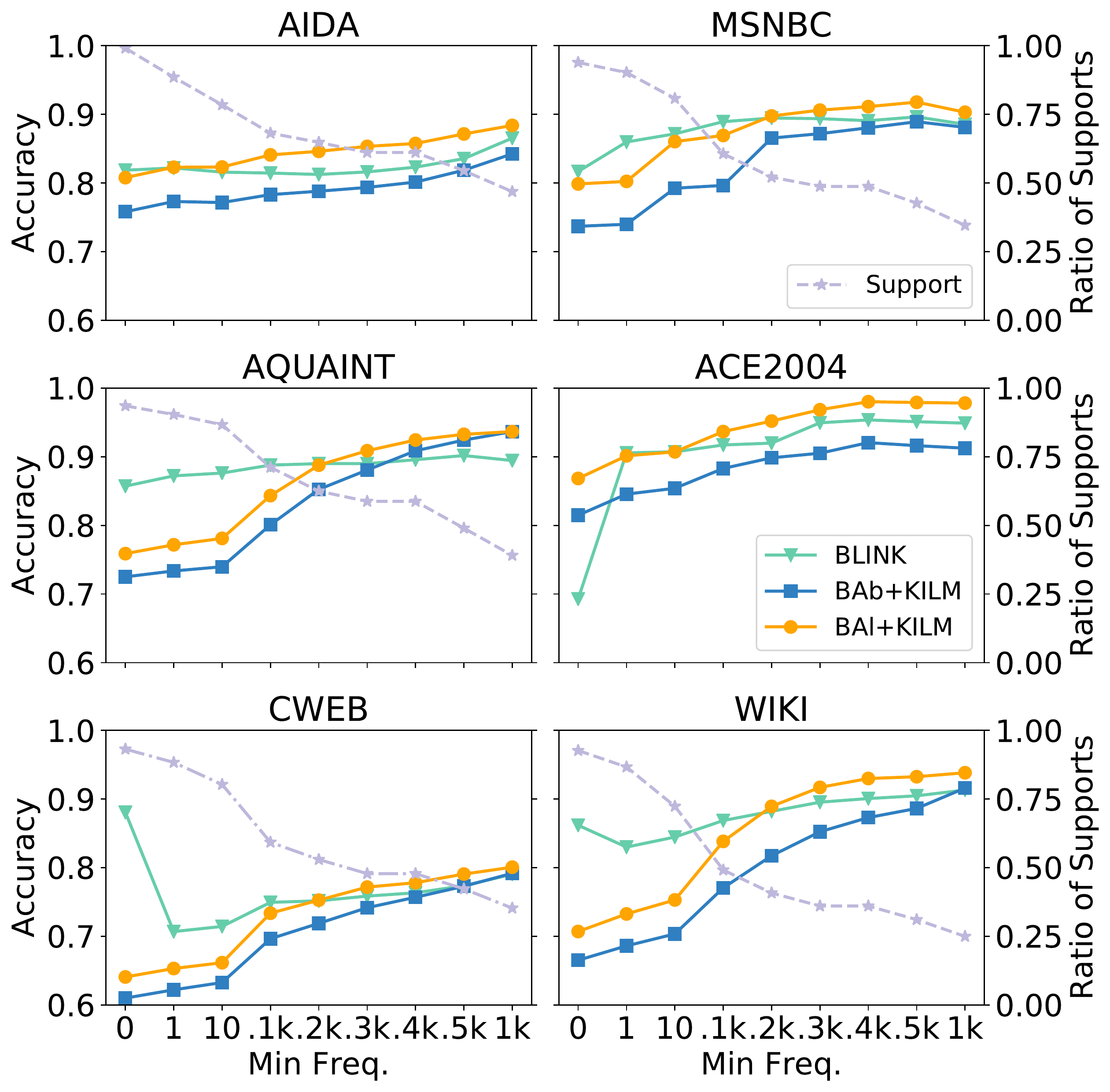} 
  \caption{Results of entity disambiguation tasks with the top five candidates and different minimum frequencies at which the target entity is sampled during the \continuedtraining. \textit{BAb} and \textit{BAl} denote BART models in base and large sizes. The primary Y-axis shows the performances of the models after KILM and the BLINK baseline on accuracy, while the Y-axis on the right shows the number of data samples that satisfy each setting.}
  \label{fig:blink}
\end{figure}

\begin{table}[t]
\centering
\resizebox{\linewidth}{!}{
\begin{tabular}[t]{@{}lccc@{}}
\toprule
\textbf{Model} & \textbf{TriviaQA} & \textbf{NQ} & \textbf{WQ} \\ \midrule
\multicolumn{4}{l}{\textit{\textbf{Finetuned settings}}} \\
RAG (Open-domain) & 68.0 & 44.5 & 45.5 \\
T5-base (Closed-Book) & 29.1 & 25.9 & 27.9 \\ \midrule
\multicolumn{4}{l}{\textit{\textbf{One/Few-shot settings}}} \\
KALM-base  & 5.87 & 1.75 & 3.53 \\
BART-base  & 9.61 & 2.19 & 3.94 \\
\quad+KILM   & \textbf{12.55} & \textbf{6.95} & \textbf{10.38} \\ \midrule
KALM-large & 11.68 & 4.34 & 6.56 \\
BART-large & 15.74 & 3.80 & 9.25 \\
\quad+KILM  & \textbf{16.42} & \textbf{7.83} & \textbf{12.65} \\ \bottomrule
\end{tabular}
}
\caption{Results on open-domain QA datasets. The best results are marked in bold. The results of the previous models except BART are taken from \citep{lewis2020retrieval,roberts2020much}.}
\label{tab:qa}
\end{table}

\begin{table*}[t!]
\centering
\resizebox{.8\linewidth}{!}{%
\begin{tabular}{@{}lcccccccccccc@{}}
\toprule
\multirow{2}{*}{Method} & \multicolumn{2}{c}{News ORG} & \multicolumn{2}{c}{News PER} & \multicolumn{2}{c}{Wiki ORG} & \multicolumn{2}{c}{Wiki PER} \\ \cmidrule(l){2-3} \cmidrule(l){4-5} \cmidrule(l){6-7} \cmidrule(l){8-9} 
 & F1 & METEOR & F1 & METEOR & F1 & METEOR & F1 & METEOR \\ \midrule
\multicolumn{9}{l}{\textit{\textbf{Constrained setting}}} \\
ApposCorpus$_{\text{constrained}}^\dagger$ & - & - & \underline{19.6} & 7.9 & - & - & - & - \\
ApposCorpus$_{\text{end2end}}^\dagger$ & - & - & 10.8 & 3.4 & - & - & - & - \\ \midrule
BART-base & 8.4 & 2.4 & \textbf{12.1} & \textbf{5.6}  & 5.2 & 1.9 & \textbf{9.2} & \textbf{4.3}  \\ 
\quad+KILM & \textbf{17.6} & \textbf{8.1} & 9.7 & 3.7 & \textbf{9.7} & \textbf{4.4} & 8.8 & 3.7 \\ \midrule
BART-large & 10.8 & 4.7 & \textbf{15.9} & \textbf{8.3} & 8.1 & 3.9 & \textbf{13.1} & \textbf{7.2} \\ 
\quad+KILM & \textbf{18.0} & \textbf{7.8} & 13.9 & 5.9 & \textbf{9.5} & \textbf{4.5} & 9.9 & 4.4 \\ \midrule \midrule
\multicolumn{9}{l}{\textit{\textbf{Non-empty setting}}} \\
BART-base & 6.6 & 2.1 & \textbf{11.7} & \textbf{4.8} & 4.4 & 1.6 & \textbf{7.5} & \textbf{3.5} \\
\quad+KILM & \textbf{14.1} & \textbf{6.7}  & 7.2 & 2.7 & \textbf{6.7} & \textbf{3.1} & 5.8 & 2.5 \\ \midrule
BART-large & 8.7 & 4.0 & \textbf{14.9} & \textbf{6.7} & \textbf{6.8} & \textbf{3.2}  & \textbf{10.6} & \textbf{6.0} \\
\quad+KILM & \textbf{14.8} & \textbf{6.6} & 9.7 & 4.1 & 6.6 & \textbf{3.2} & 6.5 & 3.0 \\\bottomrule
\end{tabular}
}
\caption{Results on zero-shot Appositive Generation under the constrained and non-empty settings. \textit{ORG} and \textit{PER} represent that the data samples are \textit{Person}- and \textit{Organization}-type entities. Bold results denote better performances of one over another with the same settings between BART and BART+KILM. $^\dagger$The results are taken from the original ApposCorpus paper, where \textit{ApposCorpus$_{\text{constrained}}$} denotes that the model is trained only with constrained data samples and \textit{ApposCorpus$_{\text{end2end}}$} denotes that the model is trained with all the data samples in a specific domain. The result hightlighted with \underline{underline} denotes that it outperforms both BART and BART+KILM.}
\label{tab:appos}
\end{table*}

\begin{table*}[t]
\centering
\resizebox{.75\linewidth}{!}{%
\begin{tabular}{@{}lccccccccc@{}}
\toprule
\multirow{2}{*}{\textbf{Model}}    & \textbf{MNLI}      & \textbf{SST} & \textbf{QQP}  & \textbf{QNLI} & \textbf{STS-B} & \textbf{RTE}  & \textbf{MRPC} & \textbf{CoLA} & \textbf{Avg} \\
                          & m/mm      & Acc   & Acc  & Acc  & Acc   & Acc  & Acc  & Mcc & -  \\ \midrule
BART-base$^\dagger$       & \textbf{85.7}/\textbf{85.8} & \textbf{93.7} & 91.3 & \textbf{91.6} & \textbf{89.9} & 74.3 & 86.4 & 51.3 & 83.3  \\
\quad+KILM                & \textbf{85.7}/85.6 & 93.0 & \textbf{91.4} & \textbf{91.6} & 89.8 & \textbf{74.9} & \textbf{87.8} & \textbf{54.2} & \textbf{83.8}  \\ \midrule
BART-large$^\dagger$      & \textbf{90.0}$^*$/\textbf{90.0} & \textbf{96.4} & 92.2 & \textbf{94.8} & \textbf{91.7}$^*$ & 82.3 & 89.5 & 57.1 & 87.1  \\
\quad+KILM                & 89.5/89.8 & 96.2 & \textbf{92.3}$^*$ & 94.7 & 91.3 & \textbf{87.0}$^*$ & \textbf{89.6} & \textbf{58.7} & \textbf{87.7}  \\ \bottomrule
\end{tabular}
}
\caption{Results on the GLUE benchmark. We report accuracy for the first seven tasks, the Matthews correlation for the CoLA dataset, and the average score (Avg) over all the tasks. $^{*}p<0.05$ with pairwise t-test.}
\label{tab:glue}
\end{table*}

\begin{table}[t]
\centering
\resizebox{\linewidth}{!}{%
\begin{tabular}{@{}lcccccc@{}}
\toprule
\multirow{2}{*}{Model}    & \multicolumn{3}{c}{\textbf{CNN Dailymail}}     & \multicolumn{3}{c}{\textbf{XSUM}}   \\ \cmidrule(l){2-4} \cmidrule(l){5-7}
                          & R1 & R2 & RL            & R1 & R2 & RL            \\ \midrule
BART-base$^\dagger$    & 42.79 & \textbf{20.31} & 39.93 & \textbf{40.83}$^*$ & \textbf{18.18}$^*$ & \textbf{33.12}$^*$ \\
\quad+KILM             & \textbf{42.86} & 20.24 & \textbf{39.94} & 40.76 & 18.15 & 33.09 \\ \midrule
BART-large$^\dagger$   & \textbf{44.14}$^*$ & \textbf{21.43}$^*$ & \textbf{41.24}$^*$ & \textbf{45.17} & \textbf{22.10} & \textbf{37.06} \\
\quad+KILM             & 43.15 & 20.86 & 40.36 & 45.07 & 21.93 & 36.95  \\\bottomrule
\end{tabular}
}
\caption{Results on summarization datasets, evaluating with ROUGE metrics. $^\dagger$The results of the BART models are re-run with the original settings except maximum sequence length to be 1024. $^{*}p<0.05$ with pairwise t-test.}
\label{tab:summ_full}
\end{table}

\section{Additional Details for Experiments}

\subsection{Pre-training Settings}
\label{sec:appx:pretrain}
We initialize the model with the original BART weights and it is continually trained on eight V100 GPUs with a batch size of 8,192. The models are optimized by the Adam optimizer with a linear scheduler and weight decay as 0.01. The peak learning rate is $5e-5$. Moreover, the maximum text length of the sequences with a knowledge component is set as 640. The mask probability and the hyper-parameter $\lambda$ for Poisson distribution are the same as those of BART. The implementation is mainly based on HuggingFace Transformers~\citep{wolf-etal-2020-transformers} and Datasets~\citep{lhoest-etal-2021-datasets} packages.

It is worth mentioning that more than 2.3 million entities with short descriptions are involved in the pre-training, and, needless to say, the occurrence of entities in Wikipedia articles is not equally distributed. For instance, while only 2,526 entities appear more than 1,000 times in the primary setting, 40.5\% of the entities only appear once in the training corpus.

\begin{table*}[h]
\centering
\resizebox{.95\linewidth}{!}{%
\begin{tabular}{@{}l|l|l@{}}
\hline
\textbf{Model} & \multicolumn{1}{c}{\textbf{Source}} & \multicolumn{1}{c}{\textbf{Input/Output Format}} \\ \hline
\multirow{4}{*}{\begin{tabular}[c]{@{}l@{}}BART+KILM\\ (ours)\end{tabular}} & \multirow{11}{*}{\begin{tabular}[c]{@{}l@{}}\textbf{\textit{Article with Entities:}} \\ The Joker is a comic book series published \\ by {[}{[}DC Comics{]}{]} starring the supervillain \\ the {[}{[}Joker{]}{]}. It ran for nine issues from \\ May–June 1975 to Sep.–Oct. 1976.\\ \\ \textbf{\textit{Entities \& Short Descriptions:}}\\ \textbf{DC Comics, Inc.:} American comic book \\ publisher and the flagship unit of DC \\ Entertainment, a subsidiary of Warner Bros. \\ Discovery.\\ \textbf{Joker (character):} fictional character \\ throughout the DC Universe.\end{tabular}} & \multicolumn{1}{c}{\textit{\textbf{Sample 1}}} \\
 &  & \begin{tabular}[c]{@{}l@{}}\textbf{Input:} The Joker \textless{}mask\textgreater book series published by \textless{}/ent\textgreater \\ DC Comics \textless{}/ent\textgreater{}\textless{}ent\_desc\textgreater{}\textless{}mask\textgreater{}\textless{}/ent\_desc\textgreater starring\\  the \textless{}mask\textgreater the Joker. It ran for nine issues from May–June \\ 1975 to Sep \textless{}mask\textgreater{}.\\ \textbf{Output:} The Joker is a comic book series published by DC \\ Comics\textless{}/ent\textgreater{}\textless{}ent\_desc\textgreater DC Comics, Inc. \textless{}sep\textgreater \\ American comic book publisher and the flagship unit of DC \\ Entertainment, a subsidiary of Warner Bros. Discovery. \\ \textless{}/ent\_desc\textgreater{}. starring the supervillain the Joker  It ran for \\ nine issues from May–June 1975 to Sep.–Oct. 1976.\end{tabular} \\
 &  & \multicolumn{1}{c}{\textit{\textbf{Sample 2}}} \\
 &  & \begin{tabular}[c]{@{}l@{}}\textbf{Input:} The Joker is a comic \textless{}mask\textgreater by DC Comics starring \\ \textless{}mask\textgreater supervillain the \textless{}ent\textgreater Joker \textless{}/ent\textgreater{}\textless{}ent\_desc\textgreater\\ \textless{}mask\textgreater{}\textless{}/ent\_desc\textgreater{}. It ran for nine issues from May \textless{}mask\textgreater \\ Sep. – Oct. 1976.\\ \textbf{Output:} The Joker is a comic book series published by DC \\ Comics starring the supervillain the \textless{}ent\textgreater Joker \textless{}/ent\textgreater\\ \textless{}ent\_desc\textgreater Joker (character) \textless{}sep\textgreater fictional character \\ throughout the DC Universe \textless{}/ent\_desc\textgreater{}. It ran for nine \\ issues from May–June 1975 to Sep.–Oct. 1976.\end{tabular} \\ \cline{1-1} \cline{3-3} 
\multirow{6}{*}{\begin{tabular}[c]{@{}l@{}}BART+Merge\\ (baseline)\end{tabular}} &  & \multicolumn{1}{c}{\textit{\textbf{Sample 1}}} \\
 &  & \begin{tabular}[c]{@{}l@{}}\textbf{Input:} The Joker \textless{}mask\textgreater book series published by DC \\ Comics starring the \textless{}mask\textgreater the Joker. It ran for nine issues \\ from May–June 1975 to Sep \textless{}mask\textgreater{}. \\ \textbf{Output:} The Joker is a comic book series published by DC \\ Comics. starring the supervillain the Joker.  It ran for nine \\ issues from May–June 1975 to Sep.–Oct. 1976.\end{tabular} \\
 &  & \multicolumn{1}{c}{\textit{\textbf{Sample 2}}} \\
 &  & \begin{tabular}[c]{@{}l@{}}\textbf{Input:} DC Comics, Inc. is American \textless{}mask\textgreater and the flag-\\ ship unit of DC \textless{}mask\textgreater{}, a subsidiary of \textless{}mask\textgreater Discovery.\\ \textbf{Output:} DC Comics, Inc. is American comic book publisher \\ and the flagship unit of DC Entertainment, a subsidiary of \\ Warner Bros. Discovery.\end{tabular} \\
 &  & \multicolumn{1}{c}{\textit{\textbf{Sample 3}}} \\
 &  & \begin{tabular}[c]{@{}l@{}}\textbf{Input:} Joker \textless{}mask\textgreater fictional character \textless{}mask\textgreater Universe.\\ \textbf{Output:} Joker (character) is fictional character throughout \\ the DC Universe.\end{tabular} \\ \cline{1-1} \cline{3-3} 
Original BART &  & \begin{tabular}[c]{@{}l@{}}\textbf{Input:} It ran for nine issues from May \textless{}mask\textgreater Sep. – \\ Oct. 1976. The Joker is a comic \textless{}mask\textgreater by DC Comics \\ starring \textless{}mask\textgreater supervillain the Joker.\\ \textbf{Output:} The Joker is a comic book series published by DC \\ Comics starring the supervillain the Joker. It ran for nine \\ issues from May–June 1975 to Sep.–Oct. 1976.\end{tabular} \\ \hline

\end{tabular}
}
\caption{Demonstrations of input and output formats of the pre-trained models involved in this work. ``BART+KILM'' denotes the models that are continued pre-trained with our proposed method; ``BART+Merge'' denotes the situation when BART model is continued pre-trained on a merge of Wikipedia corpus and the entity short descriptions; ``BART'' row shows the input and output formats of the original pre-training process of BART models.}
\label{tab:io}
\end{table*}

\subsection{Pre-training Format}
We use a piece of Wikipedia article to demonstrate the input and output formats of the involved pre-trained models involved in \Cref{tab:io}.

\subsection{Zero-shot Entity Disambiguation}
\label{sec:ed2}
As shown in ~\Cref{sec:ed}, we include the performance of BART and BART+Merge for reference. Due to the lack of conventional methods for evaluating BART models on zero-shot entity disambiguation tasks, we are inspired by the entity disambiguation model BLINK~\citep{wu2020scalable}. We evaluate BART and BART+Merge by selecting the lowest perplexity candidate that generates the corresponding Wikipedia summary/short description from a given context.
In addition, we also use the same datasets and the candidate sets as those in BLINK for more experiments. The InKB micro-F1 results are shown in~\Cref{tab:el2}, where BLINK is an entity linking model trained on TACKBP-2010 dataset.
BLINK outperforms BART+KILM in the primary setting in all but one of the datasets, but BART+KILM$_{\text{DU}}$ in data upscaling setting largely closes the performance gap between BLINK. It should be noted that both BART+KILM is a general PLM, while BLINK is not.

\paragraph{Entity Frequency in Pre-training Data} To study how the frequency of entities appearing in the pre-training text affects the entity linking performance, \Cref{fig:blink} also shows the results of experimenting with data samples with different minimum frequencies of sampling the target entity during KILM pre-training in the primary setting. As the minimum frequency increases, the gap between BART+KILM and BLINK reduces.

\subsection{Appositive Generation}
\label{sec:appx:appos}


We conduct zero-shot probing on ApposCorpus~\citep{kementchedjhieva2020apposcorpus}. We display the structured prompts of BART with \kilm~in \Cref{tab:examples}.
Following ApposCorpus, we use unigram F1 and METEOR~\citep{banerjee2005meteor} for evaluation. The results under constrained and non-empty settings are listed in \Cref{tab:appos}. Baseline results for \textit{Person}-type entities in \textit{News} domain come with the original ApposCorpus paper, while \textit{ApposCorpus$_{\text{constrained}}$} denotes that the model is trained only with constrained data samples and \textit{ApposCorpus$_{\text{end2end}}$} denotes that the model is trained with all the data samples in a specific domain. BART+\kilm~shows its advantage over BART for the \textit{Organization}-type entities, while BART outperforms BART+\kilm~on all other entity types. However, as seen in~\Cref{tab:human:appos}, the distinction in results between human evaluation and automatic metrics demonstrate how the latter do not capture important dimensions such as hallucinations.

\subsection{In-Context Few-Shot QA}

In \Cref{tab:qa}, we list the QA results when providing one example QA pairs into the inputs (1-shot) to BART models with and without KILM. Aligning with the QA example in \Cref{tab:examples}, the general evaluation format is as follows:

\begin{center}
\texttt{Question: \underline{Example Q} Answer: \underline{Example A}\textbackslash n Question: \underline{Test Q} Answer:} \colorbox{myblue}{\texttt{<mask>}}.
\end{center}

Besides BART, we also compare our performances with KALM~\citep{rosset2020knowledge} under an 8-shot setting, for which the eight examples are human-written, and two finetuned models with similar model sizes. Despite the performance gap with finetuned models, BART+KILM shows a significant advantage over the original model and KALM on all the datasets, especially for large-size models. The 1-shot results of BART-base+KILM are even higher than those of KALM-large, which has many more trainable parameters.

\subsection{Fine-tuning Experiments}
\label{sec:appx:ft}

For fine-tuning experiments, including GLUE, summarization, and KGRG tasks, we conduct each experiment with random seeds 0, 42, and 852. The numbers reported in \Cref{tab:glue+summ:simple}, \Cref{tab:glue}, \Cref{tab:summ_full} and \Cref{tab:wow} above are the averages of the results with three random seeds. The results of BART are re-run with the original settings except maximum sequence length to be 1024 for summarization tasks. Pairwise t-tests are conducted to verify the significance level of the results of BART+KILM over the baseline model.

\subsection{Human Evaluation}
\label{sec:appx:human}

For both appositive generation and KGRG task, we conduct human evaluation for a comprehensive study. Pairwise A/B testing is utilized to compare the performances of BART before and after KILM (in the primary setting). For each comparison, the same context and two options generated by the models for comparison are first randomly shuffled and then are shown to the annotators. Both tasks evaluate the performances on whether the generations are hallucinated or not, named \textit{Not Hallucinated (NH.)}. We also include two more factors for each task. For ApposCorpus, we also evaluate the generated appositives from \textit{Is Appositive (Ap.)} and \textit{Preference (Pref.)}, while we evaluate \textit{Fluency (Flu.)} and \textit{Informativeness (Info.)} for WoW. Because the dialogue task feature, we only consider the \textit{NH.} factor when the generated response is informative for WoW task. Pairwise A/B testing is utilized to compare the performances of BART before and after KILM on both ApposCorpus and WoW. Human evaluation is done among a group of experts fluent in English coming from countries across Asia. For each comparison, the same context and the generations from both models for comparison are shown to the annotators. The annotators are supposed to choose among ``generation A'', ``generation B'', ``both'', and ``neither''. Especially for the factor \textit{NH.}, the annotators are asked to search on the Internet for hallucination validation. Each comparison requires three judgments. We randomly sample 50 data samples from each subsets of ApposCorpus and 100 data samples from each WoW test set. Finally, 600 annotations are collected in total for both tasks.

\section{Datasets}
\label{sec:appx:data}
A number of datasets for downstream task evaluation are involved in this work:

\paragraph{GLUE Benchmark} GLUE benchmark is a collection of text classification datasets, which is widely used to evaluate the language modeling ability of large PLMs. In this benchmark, nine datasets are involved, including binary QA and NLI tasks. In this paper, we exclude WNLI~\citep{morgenstern2015winograd} task during evaluation because there are label conflicts in the dataset.\footnote{https://gluebenchmark.com/faq}

\paragraph{Summarization Datasets}
Text summarization is considered an essential NLG task, which requires the model to generate short summaries of long texts. In this paper, we test our models on two summarization datasets, CNN/DailyMail and XSUM. 
Summaries in the CNN/DailyMail tend to be more extractive, whereas XSUM contains highly abstractive summaries. 

\paragraph{Entity Disambiguation Datasets} The entity disambiguation task is a subtask of entity linking. Given an entity mention in the context, the model is expected to select the correct entity among a set of similar candidates. Following BLINK~\citep{wu2020scalable} and GENRE~\citep{de2020autoregressive}, we test our models on six entity disambiguation datasets, including AIDA-CoNLL dataset~\citep{hoffart2011robust}, MSNBC, AQUAINT, ACE2004, WNED-CWEB (CWEB)~\citep{gabrilovich2013facc1} and WNED-WIKI (WIKI)~\citep{guo2018robust}. We use the candidate sets from BLINK and GENRE respectively, where those of GENRE are originally from~\citet{le2018improving}.

\paragraph{ApposCorpus} Appositives are phrases that appear next to a named entity to provide background information~\citep{bauer2017nominal,kang2019pomo}. They help the readers understand the semantics of the named entities in the context. ApposCorpus~\citep{kementchedjhieva2020apposcorpus} is constructed as the first end-to-end dataset for the appositive generation task. The selected entities are \textit{Person} and \textit{Organization} entities from \textit{Wikipedia (Wiki)} and \textit{News} articles. Three types of appositives are included: constrained, empty, and a third type denoted as non-empty in this paper. Constrained appositive samples leverage WikiData for appositive generation, while empty appositive samples do not require the model to generate any appositives and non-empty samples require more general knowledge for the appositive generation. In this paper, since we do not conduct task-related training, we only evaluate our models on constrained and non-empty appositive samples.

\paragraph{Open-domain Question Answering Datasets} We further evaluate our models on three open-domain QA datasets to test the knowledge capacity: TriviaQA~\citep{joshi2017triviaqa}, Natural Questions (NQ)~\citep{kwiatkowski2019natural}, and Web Questions (WQ)~\citep{berant2013semantic}. 
TriviaQA collects the question-answer pairs from 14 trivia and quiz-league websites, where web pages and Wikipedia articles are matched to each question. NQ is a dataset of questions from web queries that can be answered with a span of Wikipedia articles. While NQ has two types of gold answers, we only evaluate the generations with the short gold answers. WQ consists of questions constructed with web queries and FreeBase~\citep{bollacker2008freebase}

\paragraph{Wizard of Wikipedia (WoW) dataset} WoW is a common crowd-sourcing KGRG dataset that relies on Wikipedia knowledge to augment the dialogue responses when discussing various topics. Two speakers are provided with an initial topic during the data collection to start the conversation. There are two test sets, \textit{seen test} and \textit{unseen test} set, split for evaluation, where the initial topics of the dialogue samples in seen test set appear in the training set and vice versa.

\setcounter{table}{2}
\renewcommand{\thetable}{C\arabic{table}}
\begin{table*}[t]
\begin{minipage}[t]{0.49\linewidth}
\centering
\resizebox{\linewidth}{!}{
\begin{tabular}{|l|l|}
\hline
\textbf{Target Entity:} & Burma Socialist Programme Party \\ \hline
\textbf{Context:} & \begin{tabular}[c]{@{}l@{}}He related the quality decline to the socialist \\ system being practiced by Burma Socialist \\ Programme Party \colorbox{pink}{<appos>} at the time. As \\ cassette players became popular, people were \\ selling their old gramophones to dealers.\end{tabular} \\ \hline
\textbf{Appostive:} &  \\
Gold & BSPP \\
BART-base & \textcolor{orange}{PSP} \\
BART-base+KILM & \begin{tabular}[c]{@{}l@{}}founded in \textcolor{orange}{1948} by Ne Win, was one of the \\ two major political parties in Burma during \\ \textcolor{orange}{the 1980–1990s}\end{tabular} \\ \hline
\end{tabular}
}
\end{minipage}
\hfill
\begin{minipage}[t]{0.49\linewidth}
\centering
\resizebox{\linewidth}{!}{
\begin{tabular}{|l|l|}
\hline
\textbf{Target Entity:} & Sultana Kamal \\ \hline
\textbf{Context:} & \begin{tabular}[c]{@{}l@{}}On April 27 , 2013 an alliance of 68 \\ non-governmental organisations working for \\ women 's rights held a rally in Dhaka to \\ speak out against the proposed subjugation \\ of women. Sultana Kamal \colorbox{pink}{<appos>} told \\ reporters there: The people are non-communal \\ and religious tolerant.\end{tabular} \\ \hline
\textbf{Appostive:} &  \\
Gold & a human rights activist \\
BART-base & \textcolor{orange}{the head of the alliance} \\
BART-base+KILM & Bangladesh \textcolor{myblue2}{politician} \\ \hline
\end{tabular}
}
\end{minipage}
\caption{Selected data samples from ApposCorpus, along with generated appositives from both BART-base and BART-base+KILM. \colorbox{pink}{\texttt{<appos>}} marks the position where an appositive is needed to enhance the target entity. In the above two tables, text in \textcolor{orange}{orange} is not factual correct, while text in  \textcolor{myblue2}{blue} is not fully accurate.}
\label{tab:example:appos}
\end{table*}

\begin{table*}[t!]
\begin{minipage}[t]{0.49\linewidth}
\centering
\resizebox{\linewidth}{!}{
\begin{tabular}{|l|l|}
\hline
\multirow{2}{*}{\textbf{\begin{tabular}[c]{@{}l@{}}Dialogue \\ History:\end{tabular}}} & \begin{tabular}[c]{@{}l@{}}B: Skiers used to only use one pole or spear \\ instead of two. Have you every gone skiing?\end{tabular} \\
 & \begin{tabular}[c]{@{}l@{}}A: I haven't ever gone skiing, but my kids \\ do and love it. Maybe one of these days I'll \\ try it out. Do you ski?\end{tabular} \\ \hline
\textbf{Response:} &  \\
BART-base & \begin{tabular}[c]{@{}l@{}}B: I do, but I don't know if I can do it. I do \\ know that \colorbox{pink}{skiing is the most popular sport} \\ \colorbox{pink}{in the world}.\end{tabular} \\
BART-base+KILM & \begin{tabular}[c]{@{}l@{}}B: I do, but I don't ski much. I'm not a big \\ fan of snow skiing.\end{tabular} \\ \hline
\end{tabular}
}
\end{minipage}
\hfill
\begin{minipage}[t]{0.49\linewidth}
\centering
\resizebox{\linewidth}{!}{
\begin{tabular}{|l|l|}
\hline
\multirow{3}{*}{\textbf{\begin{tabular}[c]{@{}l@{}}Dialogue \\ History:\end{tabular}}} & A: Nice, I walk my dog at least once a day. \\
 & \begin{tabular}[c]{@{}l@{}}B: I like watching the dog on "The Simpsons". \\ He is funny along with the family and friends.\end{tabular} \\
 & A: I love The Simpsons. \\ \hline
\textbf{Response:} &  \\
BART-base & \begin{tabular}[c]{@{}l@{}}B: The Simpsons is a great show. \colorbox{pink}{It was} \\ \colorbox{pink}{created by Homer Simpson and Matt Groening.}\end{tabular} \\
BART-base+KILM & \begin{tabular}[c]{@{}l@{}}B: I love the Simpsons too. \colorbox{pink}{It was created by} \\ \colorbox{pink}{Matt Groening.}\end{tabular} \\ \hline
\end{tabular}
}
\end{minipage}
\caption{Selected data samples from WoW, along with generated responses from both BART-base and BART-base+KILM. The highlighted generations in \colorbox{pink}{pink} is annotated to contain verifiable information.}
\label{tab:example:wow}
\end{table*}

\end{document}